\documentclass[letterpaper, 10 pt, conference]{ieeeconf}  

\IEEEoverridecommandlockouts                              
\usepackage{graphicx} 
\usepackage{amsmath} 
\usepackage{amssymb}  
\usepackage{xfrac}
\usepackage{gensymb}
\usepackage{subfig}

\title{\LARGE \bf
Design of Extra Robotic Legs for Augmenting Human Payload Capabilities by Exploiting Singularity and Torque Redistribution*
}

\author{Daniel J. Gonzalez, \IEEEmembership{Student Member,~IEEE}, and H. Harry Asada, \IEEEmembership{Member,~IEEE}$^{1}$
\thanks{*This work was supported by National Science Foundation (NSF) and the Department of Energy (DOE) of the United States of America.}
\thanks{$^{1}$The authors are with the d'Arbeloff Laboratory for Information Systems and Technology in the Department of Mechanical Engineering,
        Massachusetts Institute of Technology, Cambridge, MA 02139, USA. Email: 
        {\tt\small \{dgonz, asada\}@mit.edu}}%
}

\begin{document}

\maketitle
\thispagestyle{empty}
\pagestyle{empty}

\begin{abstract}
We present the design of a new robotic human augmentation system that will assist the operator in carrying a heavy payload, reaching and maintaining difficult postures, and ultimately better performing their job. The Extra Robotic Legs (XRL) system is worn by the operator and consists of two articulated robotic legs that move with the operator to bear a heavy payload. The design was driven by a need to increase the effectiveness of hazardous material emergency response personnel who are encumbered by their personal protective equipment (PPE). The legs will ultimately walk, climb stairs, crouch down, and crawl with the operator while eliminating all external PPE loads on the operator. The forces involved in the most extreme loading cases were analyzed to find an effective strategy for reducing actuator loads. The analysis reveals that the maximum torque is exerted during the transition from the crawling to standing mode of motion. Peak torques are significantly reduced by leveraging redundancy in force application resulting from a closed-loop kinematic chain formed by a particular posture of the XRL. The actuators, power systems, and transmission elements were designed from the results of these analyses. Using differential mechanisms to combine the inputs of multiple actuators into a single degree of freedom, the gear reductions needed to bear the heavy loads could be kept at a minimum, enabling high bandwidth force control due to the near-direct-drive transmission. A prototype was fabricated utilizing the insights gained from these analyses and initial tests indicate the feasibility of the XRL system.

Keywords: Human Augmentation, Supernumerary Robotic Limbs, Exoskeletons, Mechanism Design, Industrial Robotics
\end{abstract}


\section{Introduction}
\begin{figure}[t]
	\centering
	\includegraphics[scale=1]{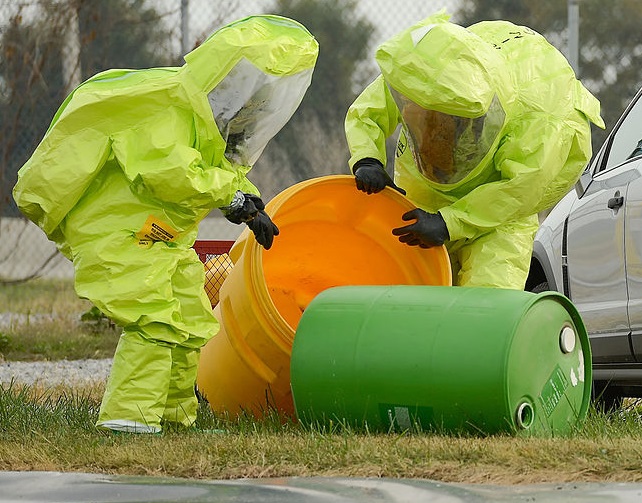}
	\caption{Hazardous Material Emergency Responders from the United States Air Force. \cite{Davis2013} Note the uncomfortable postures that must be taken while wearing the bulky and heavy hazmat suits.}
	\label{HazmatWorker}
\end{figure}

Nuclear decommissioning workers that enter areas contaminated with radioactive or other hazardous material must wear Personal Protection Equipment (PPE) such as Hazmat suits and carry Self-Contained Breathing Apparatuses (SCBA). These suits and SCBA tanks limit the amount of time a worker may stay in the working area (See Fig. \ref{HazmatWorker}). 

According to the United States Department of Energy (DOE) and their decommissioning contractors of the United Steelworkers (USW) Union, 13.6-kilogram (30-pound) Aluminum SCBA systems are currently used on site and last only 30 minutes. While 30 minutes may be enough time to arrive to the task location, perform the required task, and leave to the decontamination station, heat exhaustion usually sets in before the tank can be depleted of air\cite{NIOSH1985}. Larger and heavier air tanks, while allowing more air to be carried at a time, would only cause the worker to fatigue faster due to the effort required to carry around the added weight. 

Tethered body cooling systems have been tested by USW  at DOE facilities to little success due to unsafe exposed hoses and cables, which present a tripping and snag hazard. Self-contained body cooling systems, while addressing heat exhaustion and snag issues, would add more weight that the person must carry (5.3 kilograms, or 11.7 lbs, for the Veskimo Hydration Backpack \cite{Jechel2017}) leading to faster fatigue. 

\begin{figure}[b]
	\centering
	\includegraphics[scale=0.25]{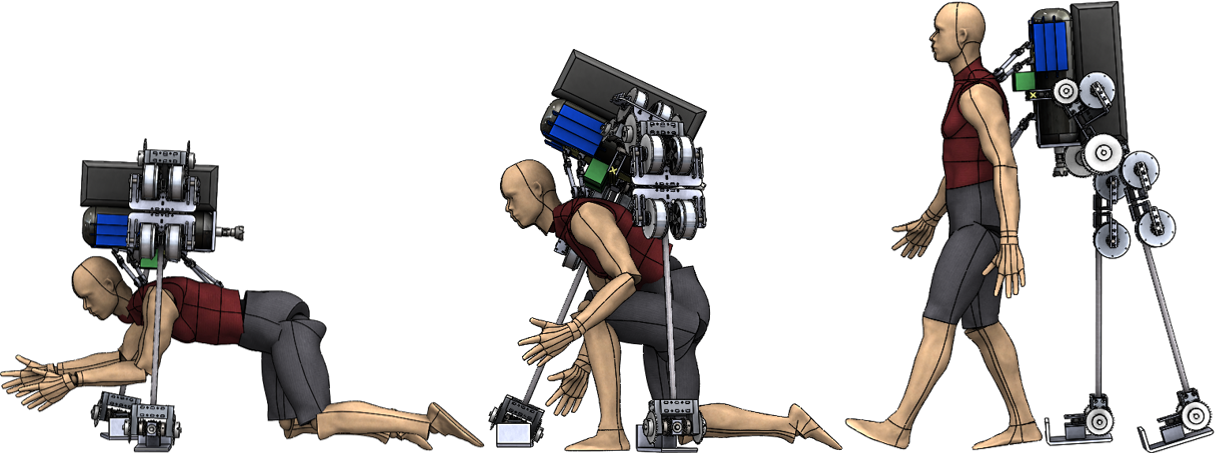}
	\caption{The Extra Robotic Legs (XRL) System}
	\label{XRLSystem}
\end{figure}
In addition to heavy loads and heat exhaustion, workers also face injury from poor body posture. Manual tasks in manufacturing and industrial settings, including decontamination work at nuclear facilities, often require workers to assume fatiguing positions near the ground. When performing these tasks, workers must kneel or crouch in potentially painful postures, sometimes using their arms to stabilize and support themselves, sharply bending at the knee and back to reach the ground and balance. These issues are compounded when workers are wearing heavy PPE. Taking such ergonomically challenging postures for long periods of time may lead to injuries in the lower back, knees, and ankles. This situation extends to a multitude of industries, including construction, manufacturing, transportation, logistics, defense, and agriculture.

According to the US Bureau of Labor Statistics, there were over 190,000 workplace injuries in manufacturing sectors, and 50,000 injuries in agriculture in 2014 \cite{BureauofLaborStatistics2015}. Overall, the cost of workplace injury amounted to over \$190 billion and resulted in over 1.1 million lost days of work \cite{Leigh2011}. Out of all workplace injuries in 2014, approximately one in three was a musculoskeletal disorder \cite{BureauofLaborStatistics2015a}. In 2010, the average civilian worker's compensation claim due to a back injury in the USA was between \$40,000 and \$80,000 \cite{CalgaryRegionalHealthAuthority2010}.

While a multitude of exoskeletons have been developed for strength augmentation \cite{Chen2016}\cite{Bogue2009}\cite{Dollar2008}, they are all tied to the kinematics of the operator, and thus limited to their range of supporting the operator. When the operator is taking kinematically unfavorable configurations, such as  squatting and crouching, the exoskeleton, too, is at a kinematically unfavorable configuration due to their similar structures. Postures that are infeasible for a person to maintain on their own, such as leaning forward while crawling on the floor with no hands, are also infeasible despite wearing an exoskeleton. They simply augment force application capabilities about a person's preexisting structure, but they do not actually compensate for the structural deficit. 

Unlike an exoskeleton, Supernumerary Robotic Limbs (SRLs) are not kinematically tied to the operator's limbs, but instead take independent structural configurations to best assist the operator. The operator's body can be braced with the robotic limbs, such as when the human is crouching. Preliminary work on stably supporting the human body with Supernumerary Limbs has been performed for both standing \cite{Parietti2015} \cite{Parietti2016} and crawling \cite{Kurek2017} postures. Our proposed solution aims to build upon these prior works by combining their features into a system that can transition seamlessly between the various augmentation modes. 

The Extra Robotic Legs (XRL) system (See Fig. \ref{XRLSystem}) allows the full weight of the SCBA and other PPE worn at the back to be borne by two autonomous articulated robotic legs. The legs walk, climb stairs, crouch down, and crawl with the operator while bearing the entire payload, eliminating all external PPE loads on the operator. In this paper, we a) explain the specific functional requirements of an augmentation solution to the load-bearing and positioning problem and introduce the XRL design concept; b) introduce a strategy for motion in the worst loading conditions that leverages the closed-loop kinematic chain present in certain configurations; c) perform kinematic and force analyses of the worst loading conditions to choose the proper actuation, power electronics, and transmission elements; d) explain the details of implementation, including the mechanical design informed by our analyses and prototype fabrication; and e) lay out our plan for future work on the XRL System. 


\section{Extra Robotic Legs for Human Augmentation}\label{FRDP}
\subsection{Functional Requirements}
The design of the XRL system was driven by the following specific functional requirements:
\begin{itemize}
	\item Must stand and balance while bearing a 22.7 kilogram (50 lbs) payload.
	\item Must crouch while providing 222.4 Newtons (50 lbs-force) of assistive lifting force to operator torso and bearing the 22.7 kilogram (50 pound) payload.
	\item Must transition between standing and crouching while providing assistive load to the operator.
 \item Must walk while standing and bearing the 22.7 kilogram (50 lbs) payload.
 \item Must crawl on the ground with the operator while providing 222.4 Newtons (50 lbs-force) of assistive lifting force to operator torso and bearing the 22.7 kilogram (50 pound) payload.
 \item Must climb and descend stairs while bearing the 22.7 kilogram (50 lbs) payload.
 \item Must last 1 hour during expected use. 
\end{itemize}

\subsection{Design Concept}
\begin{figure}[ht]
	\centering
	\includegraphics[scale=0.5]{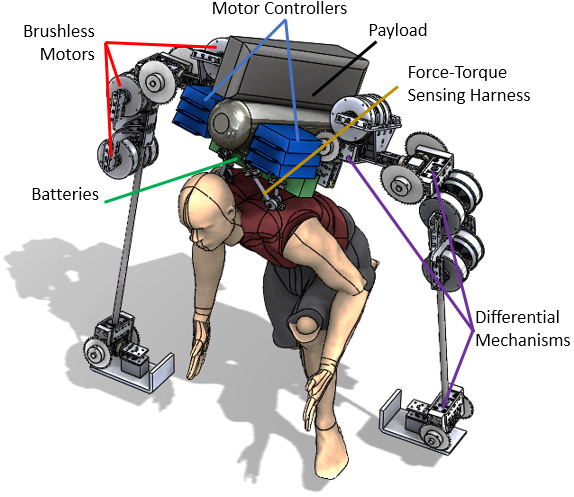}
	\caption{Layout of the modules of the XRL System}
	\label{LayoutFig}
\end{figure}

The Extra Robotic Legs (XRL) system (See Fig. \ref{LayoutFig}) consists of two fully-articulated (6-DOF) robotic legs worn at the back, which bear a payload mounted between the robot's hips. The robot is attached to the person wearing a Yates 380 full-body rappel harness via a 6-DOF force-torque sensing interface \cite{Ruiz2017} developed in the authors' laboratory.

The challenge is load bearing capacity. Merely for increasing actuator torque, use of high gear-ratio actuators may suffice. However, for stable walking and crawling the legs must move quickly. Furthermore, to better control the dynamic interactions with the ground, backdrivable actuators with high torque and low gear reduction are desirable, but are limited in load bearing capacity. To overcome these conflicting requirements, the current work explores two strategies. One is to exploit kinematic singularity, or near singular behavior where the mechanical advantage significantly increases. The other is to exploit a closed-loop kinematic chain where internal force and torque can be arbitrarily controlled so that the overall torque requirements can be lowered for individual actuators.

\subsection{Design for Near Singular Operations}
The design begins with the linkage geometry of the robot in order to ensure energy efficient operation during the most common usage situations, namely while upright (standing or walking) and while crawling. If the legs are positioned such that the motors do not need to apply any torque to bear the assistive and payload forces, then the structure itself will bear all of those forces. 
We can choose the linkage geometry such that the robot tends to be near kinematic singularity for the majority of situations during use, leading to longer and more efficient operation.

\begin{figure}[ht]
	\centering
	\includegraphics[scale=0.35]{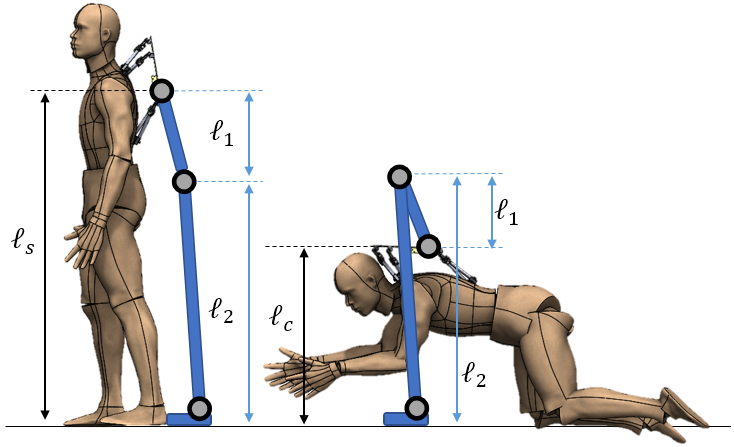}
	\caption{Geometric Design Parameters for link length and relevant geometry. The link lengths were chosen such that configurations where the human-robot system must operate for extended periods of time would be the most energy efficient. The structure bears most of the weight at near-singular configurations.}
	\label{LengthConstraintsFig}
\end{figure}
Fig. \ref{LengthConstraintsFig} demonstrates the two most common situations during operation: crawling and standing. The leg attachment point on the robot in each of these situations has a different vertical height, denoted by $\ell_s$ while standing and $\ell_c$ when crawling. The first and second links of the robot’s leg are denoted by $\ell_1$ and $\ell_2$, respectively. The robot leg link lengths were chosen to meet the following conditions:
\begin{equation}\label{length1}
\ell_s = \ell_1 + \ell_2
\end{equation}
\begin{equation}\label{length2}
\ell_c = \ell_2 - \ell_1
\end{equation}

Note that the specific link lengths are parametrized by the operator's own shoulder height and crawling height. By specifying an $\ell_s$ and $\ell_c$, measured empirically from the operator's own dimensions, the simultaneous equations \ref{length1} and \ref{length2} can be solved to find an $\ell_1$ and $\ell_2$ that will satisfy the conditions.

\section{Exploiting Closed-Loop Kinematics} \label{DPA}
One extreme case of force exertion occurs when the operator is transitioning between the crawling and standing configurations. While each posture requires little to no actuation to maintain, due to the proximity to kinematic singularities in which gravitational loads are borne only by the robot's structure, the robot must pass through a kinematically unfavorable position to transition from one configuration to the other.

The robot must carry itself, the payload, and provide 222.4 Newtons (50 lbs-force) of assistance to the operator while standing up. The total gravitational load the system must compensate for is the sum of the robot's own estimated mass of 36.3 kilograms (80 lbs), the payload mass of 22.7 kilograms (50 lbs), and the additional 222.4 Newton (50 pound-force) assistive load, for a total gravitational load of 800.7 Newtons (180 lbs-force).

\begin{figure}[ht]
	\centering
	\includegraphics[scale=0.25]{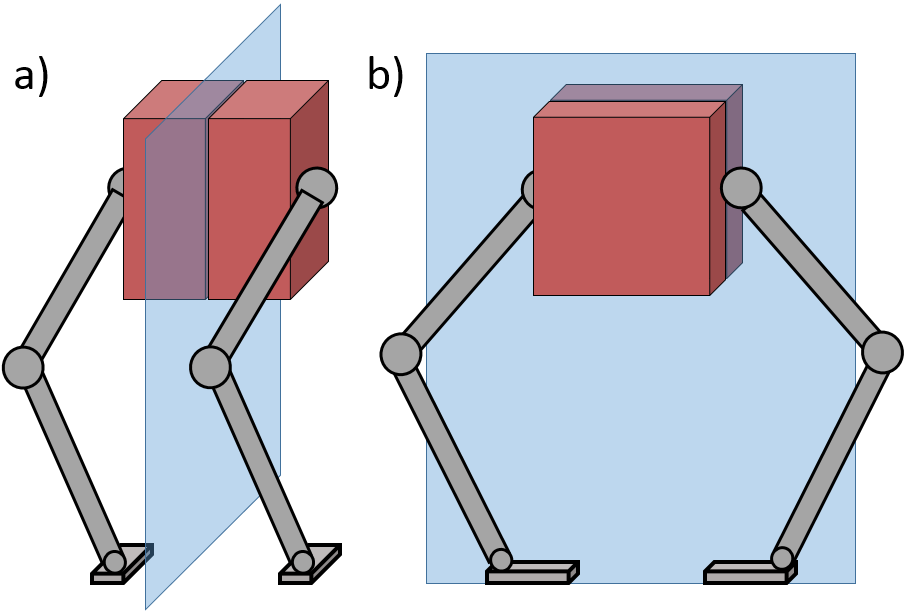}
	\caption{Rear view of the robot, facing into the page, in two knee configurations. The first a) is the sagittal (side view) plane mode where its knees are facing back. The second b) is the frontal plane mode where its knees are facing outward.}
	\label{PlaneViewFig}
\end{figure}
During the transition between the crawling and standing postures, the robot must go through a transitory squat posture. Note that the robot can take diverse configurations in order to make this transition. In particular, the two configurations shown in Fig. \ref{PlaneViewFig} are worth comparison. For standard humanoids and legged robots the configuration shown in a) where both legs move in the sagittal plane is a natural posture for supporting the body. In this case, each leg bears the same load. 

In contrast, the configuration b) where both legs are within the same frontal plane is not a natural posture, yet it has an important advantage in reducing the actuator torques. Note that the two legs, the ground, and the robot body form a closed-loop kinematic chain in the case of the frontal plane squat. The internal force and moment generated along the kinematic chain do not influence the static balance of the system, yet may change the load distribution among the actuators. This may reduce the required torque for some joints and redistribute it to other joints, so that the overall torque requirement may be lowered. We analyze and compare these two loading cases in the following subsections.

\subsection{Sagittal Plane Squat}
\begin{figure}[ht]
	\centering
	\includegraphics[scale=0.5]{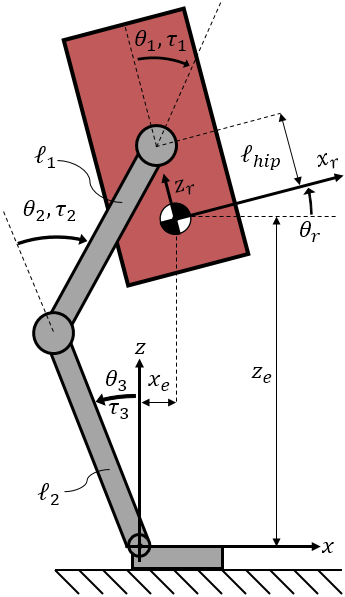}
	\caption{Geometry and Coordinates associated with the sagittal (side view) Plane Model}
	\label{SagModelFig}
\end{figure}

We consider all squatting motions to be quasi-static, and thus have no speed requirement for this motion. With the linkage geometry already specified, kinematic and force analyses may be performed for squatting motions in the sagittal (side view) plane (See Fig. \ref{SagModelFig}), when the knees are pointed back in order to prevent collisions with the operator. Note that in this image, the robot is facing to the right, in the $x_r$ direction, with the $z_r$ vector pointing upwards along the robot's torso, and pitching rotation up from the horizontal $\theta_r$ defined about the $-y_r$ direction, which faces out of the image.

Assuming symmetry about the sagittal plane, the left and right legs bear the same load. The joint torques of each leg $\left[\tau_1, \tau_2, \tau_3\right]$ required to bear external force and moment $\left[F_x, F_z, M_y\right]$ are given  
\begin{equation}\label{theJac}
\begin{bmatrix}
\tau_1\\
\tau_2\\
\tau_3  
\end{bmatrix}=\mathbb{J}^T
\begin{bmatrix}
F_x\\
F_z\\
M_y  
\end{bmatrix}
\end{equation}
where $\mathbb{J}$ is the Jacobian matrix relating endpoint velocity to joint velocities. 

The gravitational load must be borne by the two legs equally; otherwise one leg must bear a larger load, needing larger actuators. Therefore, the joint torques for performing a Sagittal Plane Squat are given by
\begin{equation}\label{theSagTorques}
\tau_{sag}=\begin{bmatrix}
\tau_{s1}\\
\tau_{s2}\\
\tau_{s3}  
\end{bmatrix}=\mathbb{J}^T
\begin{bmatrix}
0\\
-\frac{mg}{2}\\
0  
\end{bmatrix}
\end{equation}
\subsection{Frontal Plane Squat}
\begin{figure}[ht]
	\centering
	\includegraphics[scale=0.5]{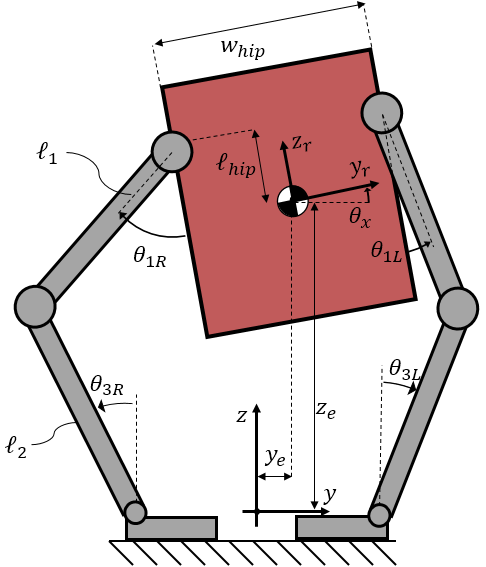}
	\caption{Geometry and Coordinates associated with the Frontal Plane Model}
	\label{PlanarModelFig}
\end{figure}

The robot's knees, which are normally pointed backward to clear the operator's body while walking, can be pointed outward to take on a configuration shown in Fig. \ref{PlanarModelFig}. Note that in this image, the robot is facing out of the page, with its left leg on the right side of the page, and its right leg on the left side of the page.

Consider the frontal plane configuration in Fig. \ref{PlanarModelFig}. We can obtain similar Jacobians for the left and right legs within the frontal plane. However, the closed-loop kinematic chain can generate internal force and moment that do not influence the static balance. Assuming symmetry again, this is achieved if the horizontal forces $F_{Right}y$ and $F_{Left}y$ sum to zero, have the same magnitude of forces in the opposite directions, and the moments $M_{Right}$ and $M_{Left}$ sum to zero. Namely,
\begin{equation}
\begin{bmatrix}
F_{Ry}\\
F_{Rz}\\
M_R 
\end{bmatrix} = 
\begin{bmatrix}
F_{y}\\
-\frac{mg}{2}\\
M
\end{bmatrix}, 
\begin{bmatrix}
F_{Ly}\\
F_{Lz}\\
M_L
\end{bmatrix} = 
\begin{bmatrix}
-F_{y}\\
-\frac{mg}{2}\\
-M
\end{bmatrix}
\end{equation}
Note that the force $F_y$ and moment $M$ are free parameters to choose for optimization. Due to the symmetricity, the squared joint torque is the same for both right and left legs. 

\subsection{Comparison and Optimization}
Let us consider the sum of squared torques at the three joints as the metric for optimization.
\begin{equation}
V\left(F_y, M\right)=\tau^T\tau
\end{equation}
Note that for Frontal Plane Squat, the total torque is a function of the two free internal variables, $F_y$ and moment $M$.  

The joint torque vector can be written for the right leg as
\begin{equation}
\tau=\mathbb{J}^T_{y\theta}\begin{bmatrix}
F_{y}\\
M
\end{bmatrix} - \mathbb{J}^T_{z}\frac{mg}{2}
\end{equation}
where
\begin{equation}
\mathbb{J} = \begin{bmatrix}
\mathbb{J}_{y}\\
\mathbb{J}_{z}\\
\mathbb{J}_{\theta}
\end{bmatrix}, \mathbb{J}_{y\theta} = \begin{bmatrix}
\mathbb{J}_{y}\\
\mathbb{J}_{\theta}
\end{bmatrix}\in \mathfrak{R}^{2\times3}
\end{equation}

For Sagittal Plane Squat, the joint torques can be written as 
\begin{equation}
\tau_{sag}=-\mathbb{J}^T_{z}\frac{mg}{2}
\end{equation}
Therefore, the sagittal plane joint torques are a special case of the above frontal plane joint torques where the free parameters are zero: $F_y=0$, $M_x=0$.

For Frontal Plane Squat, the minimum total torque can be obtained for the following internal force and moment:

\begin{equation}
\begin{bmatrix}
F_{y}\\
M
\end{bmatrix} 
=-\mathbb{J}^\#_{y\theta}\tau_{sag}
\end{equation}
where $\mathbb{J}^\#_{y\theta}$  is the pseudoinverse of the matrix. Therefore, the optimal joint torques can be given by
\begin{equation}
\tau^0=\left(\mathbb{I}-\mathbb{J}^T_{y\theta}\mathbb{J}^{T\#}_{y\theta}\right)\tau_{sag}
\end{equation}
With this we can show $|\tau^0|\leq|\tau_{sag}|$. The frontal plane joint torques are in general smaller than that of the sagittal plane joint torques. The numerical analysis below confirms this result.

By analyzing the case where external forces and torques $F_{z}=-mg$, $F_{y}=0$ and $M_{x}=0$, and by assuming torque output symmetry ($\tau_{Left1}=\tau_{Right1}$, etc), geometric symmetry about the z-axis, and that all torques are about the x-axis, we obtain the following simultaneous equations:
\begin{equation}\label{simul1}
-\ell_1\sin{\theta_{1}}\frac{mg}{2}=\tau_{2}-\tau_{1}
\end{equation}
\begin{equation}\label{simul2}
\ell_2\sin{\theta_{3}}\frac{mg}{2}=\tau_{3}-\tau_{2}
\end{equation}

\begin{figure}[ht]
	\centering
	\includegraphics[scale=0.8]{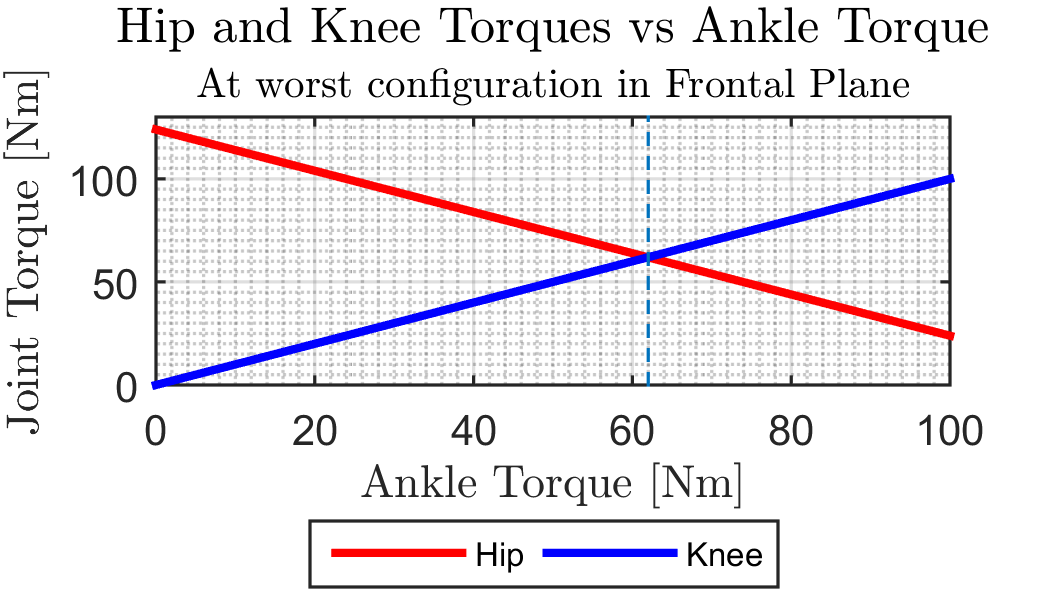}
	\caption{Redistributing the required Hip and Knee torques at a height of 1 meter by altering the Ankle torque using the redundancy offered in the Frontal Plane. Note that for this configuration, the optimal set of torques is 62 Nm for each joint.}
	\label{AlphaFig}
\end{figure}

Unlike the sagittal plane case, where a certain configuration of the open kinematic chain garnered only one solution to the joint torques required to bear the external load, multiple combinations of joint torque solutions exist to solve \eqref{simul1} and \eqref{simul2}. Because we wish to minimize the maximum torque being applied by any one motor, this is a case of Minimax optimization. (see Fig. \ref{AlphaFig})

\begin{figure}[ht]
	\centering
	\includegraphics[scale=0.8]{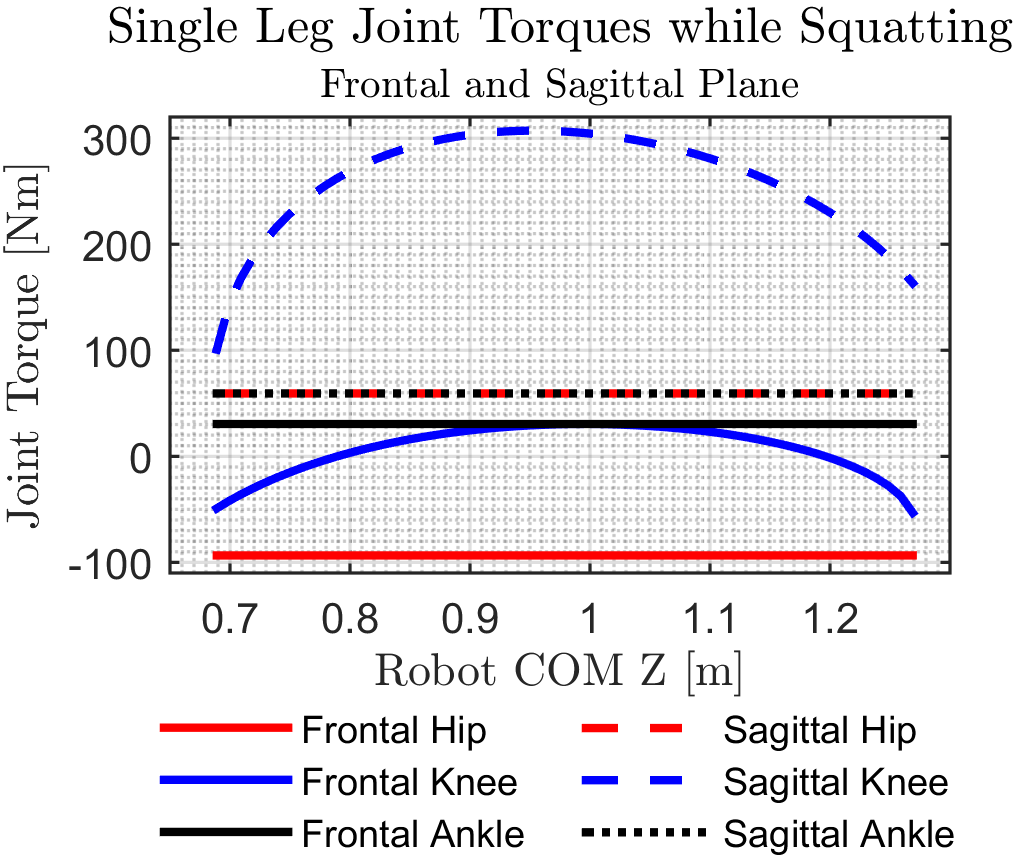}
	\caption{Comparison of joint torques in each leg required to squat with 222.4 Newton (50 pound) assistive load with knees in both the Sagittal Plane and the Frontal Plane when standing with center of mass (COM) centered between the ankle joints. Note that the maximum torque required to squat is minimized when redistributing torques using the redundancy offered in the Frontal Plane.}
	\label{BothSquat0Fig}
\end{figure}

The dashed and dotted lines of Fig. \ref{BothSquat0Fig} show the joint torques required for a single leg when both legs are in contact with the ground, the robot is level, standing directly over its ankle, and performing a squatting motion with its knees in the sagittal configuration. The link lengths $\ell_1$ and $\ell_2$ were chosen with the worst-case assumption that the operator is a 97.5th percentile American male with a height of 194.31 cm (6 feet 4.5 inches) \cite{Dreyfuss1960}. 

The constant 59 Nm that the hip and ankle must apply come from the assistive upward force that is being applied to the operator. The robot must also apply a rearing torque to be able to apply that assistance to the operator's center of mass. The knee sees a peak torque of 307 Nm at its most kinematically unfavorable configuration mid-squat, when link 1 is completely horizontal. Designing a robot to be able to apply these loads, let alone transfer all of the load to one leg in order to lift the other and take a step, would require powerful actuators and heavy gearing. By choosing  the frontal plane strategy for squatting, we can exploit  the closed-chain kinematics to significantly reduce the actuation torque requirement when squatting and providing an assistive load.

Fig. \ref{BothSquat0Fig} also shows the joint torques required in a single leg for the robot to squat when it is standing centered and level, with its knees in the frontal configuration and its ankles shoulder-width apart. In order to prevent the foot from lifting off the ground, the ankle torque is specified to be half the gravitational load times the expected width of the foot. 

The joint torques have been redistributed, with the hip and ankle applying constant torques of -93.4 Nm and 30.5 Nm, respectively, to offset the torque required by the knee to a peak of 30.5 Nm. Note that the magnitudes of these joint torques are less than the peak torque required by the knee in the sagittal plane configuration, confirming our analytical insight. 

\subsection{Stair Climbing Analysis}\label{stairanal}
The other most extreme case of force exertion occurs when the operator is climbing a set of stairs. The maximum allowable height of a stair according to the International Residential Code is 19.7 cm (7.750 inches)\cite{Council2006}. To allow for inconsistencies in stair height, a stair height of 20.3 cm (8 inches) was considered in our calculations. 

The gravitational load the robot must bear while performing these maneuvers is less than when squatting, only 578.3 Newtons (130 lbs-force), because only the robot's mass and the payload are borne while standing, and no upward assistive load to the operator is applied. The entire gravitational load must be borne by only one leg at a time, but the other leg must reach up to land on the next step. 

Because the robot will transition to climbing stairs from a standing or walking configuration, knee and ankle torques during upward and forward motion were analyzed using the sagittal plane model. Meanwhile the hip torque required to hold up the torso and flight-phase leg was analyzed using a modified frontal plane model.

Ultimately, the maximum loading conditions all occur in the condition where the robot is standing and climbing up a stair with one leg on the ground. When raising a foot and beginning to climb a step, the maximum knee torque required is 168 Nm, and the maximum ankle torque is 115.6 Nm when leaning forward 20 cm. With a hip width of 35.56 cm (14 inches), the maximum hip torque required to hold the torso up is 102.8 Nm. 
\section{Implementation}\label{DetailedDesign}
\subsection{Detailed Module Design}
In order to achieve accurate force output control and proprioceptive sensing, the XRL actuators should have as little gear reduction as possible \cite{Kenneally2016} \cite{Wensing2017} while still meeting our torque requirements. A near-direct-drive architecture can be achieved by using high torque motors and a gear reduction less than 10:1. 

\begin{figure}[ht]
	\centering
	\includegraphics[scale=0.20]{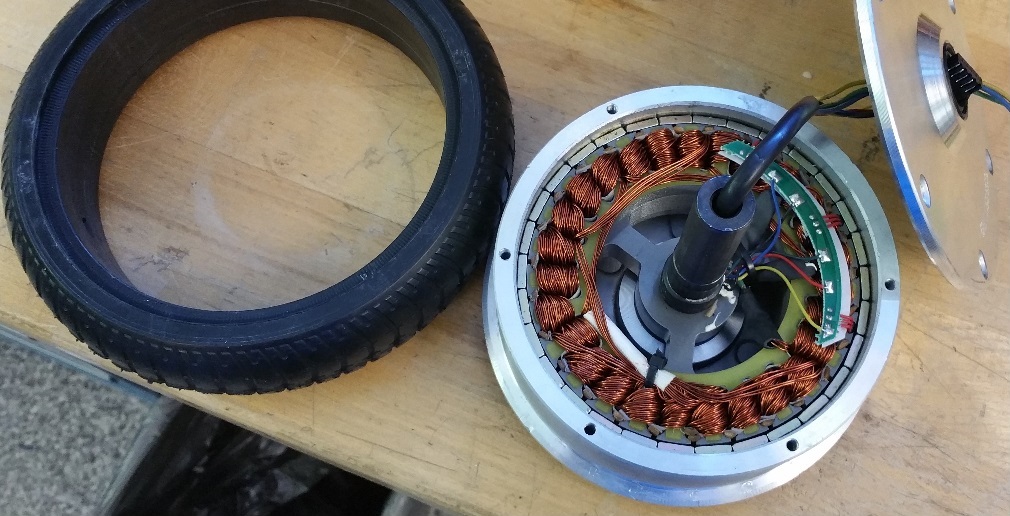}
	\caption{Disassembled high torque brushless motor from the ``Hoverboard'' self-balancing scooter.}
	\label{hoverFig}
\end{figure} 
Meeting the large torque requirements of \ref{stairanal} while keeping the gear reduction low led to two key design choices: the use of commercial high-torque brushless motors from ``Hoverboard'' self-balancing scooters (See Fig. \ref{hoverFig}), and the use of differential mechanisms at each joint. 

The 15-pole brushless outrunner ``Hoverboard'' motors are inexpensive (replacements can be purchased for 20 USD each), have built-in hall-effect sensors, and have a torque constant of 0.45 Nm/A, with little cogging torque. They have built-in deep-groove roller bearings which allow the motor to withstand large moments when cantilevered by its mounting shaft. Unfortunately, they weigh 2.2 kilograms each, but much of the weight can be reduced by removing excess material used to encapsulate the tire using a machining operation. They were verified to output 22.5 Nm at 50 A with no saturation. This motor was selected for its low price, high torque constant, and availability. Six RoboteQ HBL2360 Dual-Channel Brushless Motor Controllers, which can source 50 A continuous and 75 A peak, were used to drive and control these motors.

\begin{figure}[ht]
	\centering
	\includegraphics[scale=0.45]{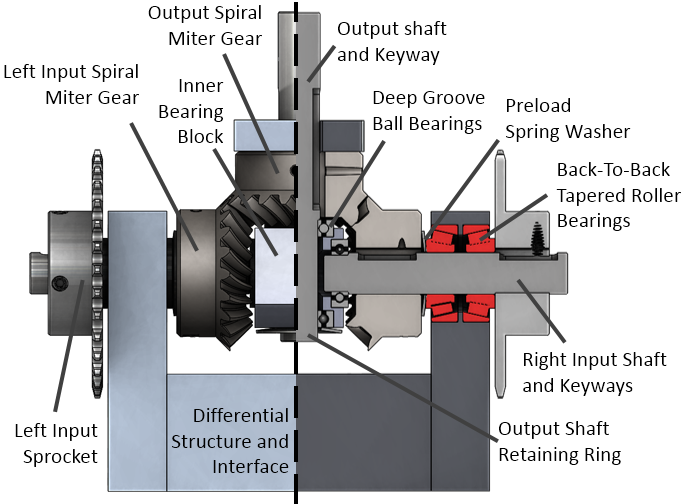}
	\caption{Cutaway solid model view of the differential mechanism used at the hip, knee, and ankle joints.}
	\label{diffFig}
\end{figure} 
A single motor at the knee would still require a 7.47:1 gear reduction, however, which would require multiple transmission stages. In order to further reduce the requirement of gearing, a differential mechanism (See Fig. \ref{diffFig}) can be used at each joint. Two actuators are connected via a chain-drive transmission to the Left and Right Input Shafts, which are coupled via KHK Spiral Miter Gears to the Output Shaft.As the two motors rotate in opposite directions, the Output Shaft coming out of the middle Spiral Miter Gear rotates about its centerline. Rotation of the two motors in the same direction generates a rotation of the entire Output Shaft about the centerline of both inputs. Rather than having one motor for each degree of freedom, both motors' combined torques may be completely directed to a single degree of freedom, doubling the effective torque.

Loads on the output shaft are first borne by the Deep Groove Roller Bearings in the Inner Bearing Block, which allows two axes of rotation. The loads are then borne by the Left and Right  Input Shafts. While each Input Shaft sees a moment and radial forces due to the cantilevered load from the Inner bearing Block, and axial loads from the Spiral Miter Gears, Tapered Roller Bearings in the back-to-back configuration can bear these loads while keeping the Input Shafts aligned. 

The result is a compact but stiff 2-DOF joint that allows the joint torque to be distributed between two motors, each requiring half of the gearing. This backdrivable design ensures the safety of the operator, as the robot will be able sense the operator through proprioception and act quickly in the case of a collision.

\subsection{Prototype}
The structural components of the differential mechanisms were fabricated on a CNC mill. The links were cut from aluminum tube, with clamping interface components made of aluminum and cut on the waterjet. The link lengths may be adjusted using these clamp mechanisms to adhere to the operator's size, and our planned kinematic and trajectory planning algorithms will adjust accordingly.

Each motor was modified by removing the rubber tire, adding a diametrically polarized magnet to the output side to measure angle with an AMS AS5147P magnetic encoder, and adding a 15-tooth \#35 chain drive sprocket to the side closest to the mounting shaft. 

\begin{figure}[ht]
	\centering
	\includegraphics[scale=0.60]{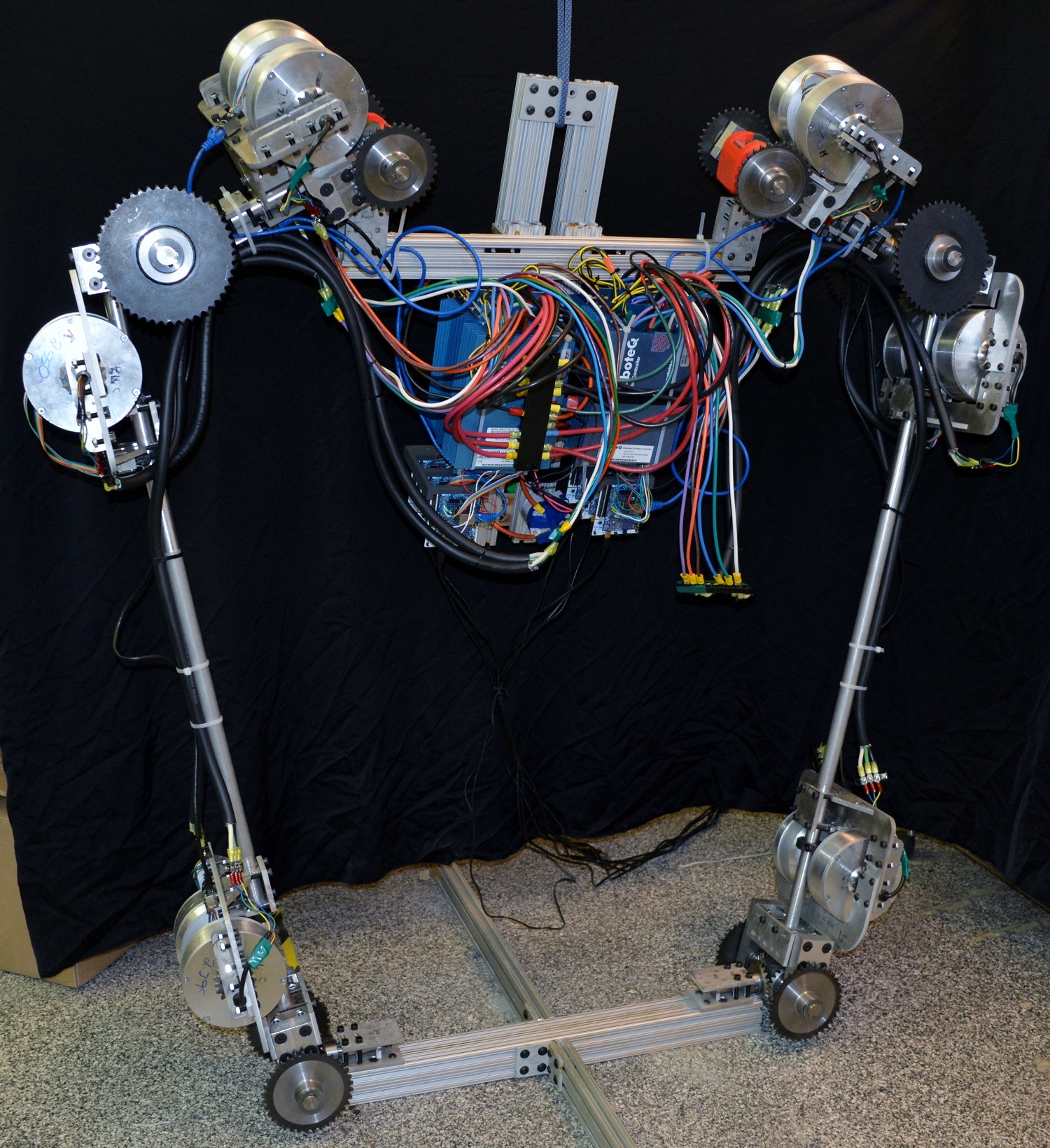}
	\caption{XRL System Prototype in the Frontal Plane squatting configuration}
	\label{protoFig}
\end{figure}
The XRL prototype (see Fig, \ref{protoFig}) uses 8020 modular aluminum extrusion to allow for rapid development during testing. The prototype will eventually attach to the Force-Torque Sensing Harness \cite{Ruiz2017} for testing with a human operator. Initial experiments were performed to verify the functionality of joint-level force and position control, as well as whole-body balance control using a high-stiffness ankle strategy. Future work includes verifying the maximum payload capacity. 

\section{Conclusion and Future Work} \label{conclude}
The XRL system augments the operator's ability to carry a heavy payload and reach and maintain difficult postures, and can supplement the current personal protective equipment worn by hazardous materials response teams and allow them to perform their job more effectively. By using two articulated legs that are independent from the operator's kinematics, we can exploit favorable configurations when the operator cannot. Assistive forces can be applied to the operator when transitioning between the crouching and standing configurations by distributing the joint torques to each actuator, eliminating the need for heavy gearing. Using differential mechanisms to combine the inputs of multiple actuators in a single degree of freedom, the gearing needed to bear the heavy loads are further reduced, resulting in a near-direct-drive actuation architecture. The resultant prototype can bear and squat with heavy loads while allowing for high-bandwidth force control.

Next steps for the XRL system include verifying the squatting strength, dynamic balance and walking control and stability, and interfacing with the operator for a seamless augmentation experience. The prototype XRL system will serve as a platform for work on intent estimation, planning, and control.

\bibliographystyle{IEEEtran}
\bibliography{IEEEabrv,myBib}
	

\end{document}